# Foundations for Understanding and Building Conscious Systems using Stable Parallel Looped Dynamics[1]

Muralidhar Ravuri[2]

## Abstract

The problem of consciousness faced several challenges for a few reasons: (a) a lack of necessary and sufficient conditions, without which we would not know how close we are to the solution, (b) a lack of a synthesis framework to build conscious systems and (c) a lack of mechanisms explaining the transition between the lower-level chemical dynamics and the higher-level abstractions. In this paper, I address these issues using a new framework. The central result is that a person is 'minimally' conscious if and only if he knows at least one truth. This lets us move away from the vagueness surrounding consciousness and instead focus equivalently on: (i) what truths are and how our brain represents/relates them to each other and (ii) how we attain a feeling of knowing for a truth. For the former problem, since truths are things that do not change, I replace the abstract notion with a dynamical one called fixed sets. These sets are guaranteed to exist for our brain and other stable parallel looped systems. The relationships between everyday events are now built using relationships between fixed sets, until our brain creates a unique dynamical state called the self-sustaining threshold 'membrane' of fixed sets. For the latter problem, I present necessary and sufficient conditions for attaining a feeling of knowing using a definition of continuity applied to abstractions. Combining these results, I now say that a person is minimally conscious if and only if his brain has a self-sustaining dynamical membrane with abstract continuous paths. A synthetic system built to satisfy this equivalent self-sustaining membrane condition appears indistinguishable from human consciousness.

---

# 1 Introduction

Life existed on earth for about 3.9 billion years. During this period, several dynamical changes have occurred, including the existence, evolution, migration and extinction of several species. Does any living or nonliving system, other than humans, know and understand such events? Among animals, we notice that some exhibit memory of a few local events and some others seem to understand even death. From these observations, most people tend to accept that human consciousness, emotions and other unique human features are present in other species as well, but perhaps to a much lesser degree. Such a gradational representation of feelings and consciousness is considered the least contradictory from a logical point of view, even though there is only a limited, if at all, scientific verification. Given such gradations, what is a suitable definition of *minimal* consciousness, i.e., the lowest degree of consciousness below which the concept has no relation to consciousness?

The typical objectives when defining minimal consciousness would be (a) to develop a rigorous theory and (b) to rigorously extend it to cover all other notions commonly attributed to human consciousness like self-awareness, free will and feelings. In this regard, for each choice of definition, we would need to rely on human consciousness and compare the differences between conscious and unconscious states. Among the unconscious or less conscious states, I list a set of seven commonly accepted ones – deep sleep, under general anesthesia, knocked unconscious, coma, severely drunk, sleep walking and even a newborn baby.

## 1.1. Necessary and sufficient conditions

There have been several attempts at providing new frameworks for consciousness (Baars 1988; Dennett 1992; Chalmers 1996; Crick & Koch 2003; Tononi 2004). Each of these recognizes individual aspects within the complexity of our brain network as the source of consciousness. A few examples of these approaches are: identifying the neural correlates for consciousness (Koch 2004), taking an information theoretic approach to consciousness by studying the connectivity in graphs (Tononi 2004), looking at the origin of consciousness within humanity (Jaynes 2000), sensorimotor account for visual consciousness (O'Regan & Noë 2001), quantum mechanical view for consciousness (Hameroff 2006) and evolutionary views on consciousness (Edelman 1987; Dennett 1992). However, none of them have attempted to specify the *necessary and sufficient* conditions for consciousness.

How can we gain confidence in a theory on consciousness unless we work with necessary and sufficient conditions? For example, Tononi (2004) has suggested that "consciousness is integrated information". Necessity of integrated information i.e., that which is beyond the sum of its parts is well justified. However, why is integrated information sufficient? What is the 'nature' of this integrated information that would make it sufficient? When we build a machine in which the information is well integrated, what are we integrating and why would we regard it as conscious? Besides, the integrated information in terms of purely static structural properties of the brain network (for example, static structural regions like fusiform gyrus in area V8 identified as neural correlate of color – Crick & Koch, 2003) cannot be sufficient. This is because the static network and regions such as fusiform gyrus are structurally the same, even if manually stimulated, both when we are conscious and when we are not (like during deep sleep and under general anesthesia). Therefore, static structures cannot be the complete set of equivalent conditions. *Dynamical properties* must necessarily be included for consciousness. Yet, there is never a mention of what physical and chemical dynamics are necessary and sufficient.

As another example, O'Regan & Noë (2001) have mentioned that visual consciousness is the outcome of sensorimotor contingencies. Here again, necessity of actions and sensing is well justified. However, why and what sensorimotor interactions would make them sufficient for consciousness? If we build a machine, how many sensorimotor contingencies should be included and in what order? When such a machine becomes complex, what is the nature of this complexity and what resulting structural and dynamical properties make it sufficient?

We can now evaluate each of the existing frameworks on consciousness in a similar way. They all fall short with either necessity or sufficiency. This is one reason why we feel we do not yet know what consciousness is.

In this paper, I deviate from existing approaches by providing both necessary and sufficient conditions for each concept introduced, even if it appears trivial at first glance. For example, the central result of this paper is the following – *minimal consciousness is equivalent to knowing at least one truth* (Claim 1). It is necessary because whenever we are conscious, we know a lot of truths (like we know our hands, legs, our emotions, tables and others). If we do not know anything like during deep sleep, surely we cannot be conscious. It is also sufficient because the moment we know a single truth, we are conscious or, equivalently, when we are not conscious (like during deep sleep, under general anesthesia and others) we do not know even a single truth (Claim 1 in Section 2).

This central result has given us one set of necessary and sufficient conditions. At first glance, this equivalence sounds either trivial or that we have pushed the problem of consciousness into the problem of knowing. The objective with the rest of the paper is to clarify and show how to build a dynamical systems framework for the equivalent conditions i.e., for both truths and knowing. Specifically, I show that a special class called stable parallel looped (SPL) dynamical systems provides the underlying framework. They help generate an infinite family of increasingly complex and highly interacting stable systems. We cannot avoid constructing such complexity when building both natural and synthetic conscious systems, as simple systems are rarely conscious. SPL systems are, for now, the only way to create systems with millions of interacting subcomponents that do not collapse for a while. Using this family, I show that *a system is minimally conscious if and only if it has a threshold self-sustaining dynamical membrane of meta-fixed sets* (Claim 8). If the membrane exists, you are minimally conscious and if it 'tears down', you are not conscious. Furthermore, Claim 8 allows us (a) to build synthetic conscious systems and (b) to study human consciousness empirically as well.

As another example for which we need necessary and sufficient conditions, consider why we understand sentences even if we have never heard them before (like sentences in this paper). Is it because of grammatical correctness? But why? Universal grammar (Chomsky 1957) is not sufficient for *understanding* language, though some representation of it appears necessary. This is because you need to be conscious first before you can understand anything. Besides, it is not clear why a machine with a universal grammar would 'understand' language. Even among humans, for example, we understand every word in 'eat to like I pizza' and, yet, not the sentence. A simple rearrangement as 'I like to eat pizza' completely changes our understanding ability. Analogously, a jumbled arrangement of individually understandable pieces of a jigsaw puzzle still does not make us understand the whole. As we start to rearrange it correctly, we do understand it. When we forgot who or where we saw a given person, we can eventually know this by recollecting and rearranging the sequence of events in a correct way. In each of these cases – correct arrangement of words, correct arrangement of jigsaw-like puzzles, correct arrangement of sound variations, correct arrangement of events and so on – what are the necessary and sufficient conditions to understand them (Sections 5-6)? I show that we need two structures – the above dynamical membrane and a notion of continuity that generalizes to abstractions (Claim 5).

## 1.2. Minimality – knowing

In the above discussion, I have picked 'knowing truths' as the minimal set for consciousness. Why should we pick them? One reason is because knowing is so fundamental that most of us treat it as synonymous to consciousness. For example, knowing yourself is self-awareness. Knowing the existence of the external world involves knowing the existence of space, knowing the passage of time, knowing the external objects and having a cohesive perception. Knowing your interactions with the external world includes your free will. Knowing your feelings is related to knowing your subjective quality of your experiences (*qualia*). Therefore, if we can formalize the concept of knowing (Fig. 1), it appears that we can extend it to the above specific forms of knowing.

Conversely, if we do not include knowing into the minimal set, the resulting definition does not appear to have any relationship to the notion of consciousness. For example, if you indeed do not know absolutely anything, why would you consider yourself conscious? This situation indeed occurs in each of the seven less conscious states mentioned above (like deep sleep, coma and under general anesthesia). *If someone says that 'nothing exists in this universe', then even though the statement appears obviously incorrect, we can guarantee that it is false **if and only if** we are conscious.* To convince ourselves, consider one of the less conscious state, namely, when we are in deep sleep. Even though I and the rest of the universe continue to exist physically, I, as a system, am incapable of 'knowing' my existence or the existence of the universe. This is true with any fact for which an adult conscious human can evaluate truth. Even if you are repeatedly performing the same task (like an newborn drinking milk every day), you do not know or realize what you were doing, why you were doing and that you were even doing it, to the same degree as adult conscious humans. This statement is at the heart of every problem related to knowing anything in the universe. This is the reason why I use it to define *minimal* consciousness. In this paper, I will only consider minimal consciousness in detail (Fig. 1). All extensions to other features of consciousness like free will and passage of time will be considered in subsequent papers.

## 1.3. Analysis versus synthesis

Another requirement for a theory of consciousness is that it should explain how to *build* synthetic conscious systems in addition to *analyzing* how our brain gives rise to consciousness. As an analogy, to fully understand what a computer is and how it works, we should also understand how to build it, at least, in theoretical terms. All existing models and discussions on consciousness either ignore or postpone this aspect. This makes them necessarily incomplete.

Towards this, specifying necessary and sufficient conditions become critical once again. Without them, there is no sure way to evaluate theoretical or experimental correctness. For example, how can we answer if neurons or brain are necessary for consciousness? It is unreasonable to suggest that there is only one approach to become conscious and that the natural evolutionary mechanisms did manage to pick it. If we are attempting to build a synthetic conscious system, when will we stop our pursuit and say that we have reached our goal if we do not have these necessary and sufficient conditions? The new theory proposed here is constructive. It is based on the stable parallel looped dynamical systems framework.

Compared to an analytical perspective, a synthesis perspective introduces a different set of questions which are equally important. I will list a few here and discuss most of them in subsequent papers. For example, imagine trying to create a system capable of knowing itself. Would the front part of the system know that the back part is part of its own body? Does a single cell, which survives as a whole entity, know parts of its own body? How about a colony of bacterial cells, a collection of cells in your brain, an entire plant or a non-living system like a robot? How could some collections become self-aware but not others? If you were to assign awareness to any of these systems, to which part would you do so – its front, its back, its middle or some combination of them. Claims Claim **5** and Claim **6** will let us evaluate such questions in precise terms for any system – living and non-living.

How does a given system *know* that there is a world outside itself? What representation of memory and internal brain structure allows this? As an adaptive system, how does our brain continuously recalibrate itself to keep memories and experiences consistent and identical (like the perception of a car remains the same even after 10 years)? As an analogy, if a sensor like a thermometer were constantly changing its internal structure like our brain, how would it recalibrate to continue measuring the same temperature? How can we move our hand quite precisely without ever knowing which motor neurons to activate and in which order (cf. a robot)? In this paper, I will address synthesis-related questions related to minimal consciousness, but postpone all others to subsequent papers.

## 1.4. Linking abstractions to dynamics

When studying consciousness, most researchers take two common and, yet, disjoint approaches. These are (a) studying lower-level dynamical details involving chemical reactions occurring within and across neurons and (b) studying higher-level abstract description of events that can be understood in a natural language or other abstract representations. In the lower-level studies, the focus is to find the neural correlates for consciousness (Koch 2004). In the higher-level studies, the description of the features like self-awareness, tables, hands and legs are at an abstract level using concepts and relationships that can only be understood by another conscious being. For now, there is a large gap between the high-level abstractions and the low-level dynamical representations. How do we connect the two representations? In this paper, I develop a one-to-one mapping with special dynamics called 'fixed sets' that let us move between the two representations seamlessly.

With abstractions like qualia, feelings and emotions, the gap is even more prominent in spite of the fact that the existence of qualia is the outcome of dynamics alone. To see this latter fact, note first that the possibility of a quale being always present (or is innate) is not true. For otherwise, we would need to attribute a quale to a dead cell or a collection of molecules obeying certain properties even outside a conscious human. Therefore, the quale is nonexistent when we were a fertilized egg cell and it comes to existence when we become an adult. During this transition, only continuous physical and chemical dynamics have occurred. In this paper, I provide necessary and sufficient conditions for one quale, namely, the feeling of knowing, as it relates to minimal consciousness (Section 6). Other qualia will be considered in subsequent papers.

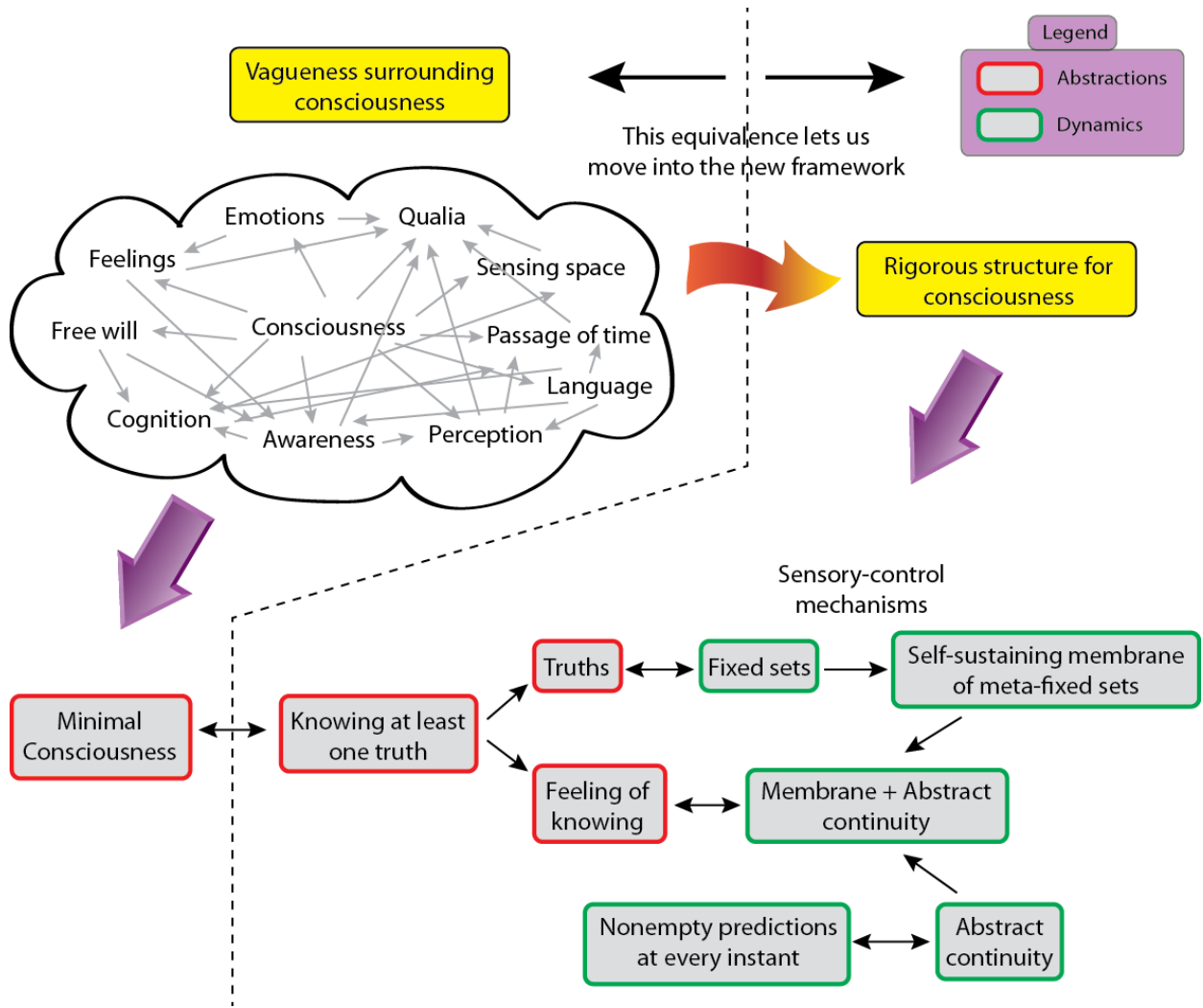

**Figure 1 | Brief outline of the paper.** The central result of the paper is that minimal consciousness is equivalent to knowing at least one truth. This result lets us move away from the intertwined nature of consciousness. Instead, it lets us study truths, the structure and relationships between them and the feeling of knowing using a dynamical framework of stable parallel loops. The resulting necessary and sufficient structure revealed by this approach is that minimal consciousness is equivalent to the existence of a self-sustaining dynamical membrane of meta-fixed sets made of abstract continuous paths.

If we now impose the synthesis requirement, what are the choices for the lower-level dynamics? Should it be based on chemical reactions and neurons or can it be based on electronic components? Identifying the necessary and sufficient conditions is once again the only hope to close this gap. In Fig. 1, I highlight the difference between abstract concepts from the dynamical ones for the sake of clarity. For example, in Section 3, the abstractions called truths are equivalently represented using a dynamical notion called fixed sets. Similarly, Claim 5 specifies necessary and sufficient conditions for attaining a feeling of knowing using a dynamical representation of abstractions. With this introduction (a brief summary in Fig. 1), let me now introduce the new theoretical framework on consciousness.

## 2. Minimal consciousness

The central result of this paper is the following claim. It provides equivalent conditions for minimal consciousness. It allows us to avoid the intricately intertwined nature of consciousness and instead pick an alternate approach (Fig. 1). This is analogous to taking an algebraic approach when solving a geometric problem.

**Claim 1:** A dynamical system is minimally conscious if and only if it knows at least one truth.

*Proof*: Since normal adult humans are the only measurable examples for consciousness, I will use them as the nonempty class of dynamical systems that validate the above claim. I will then generalize the result to all dynamical systems by treating the claim simply as a definition of a *minimally conscious* state.

   *Minimally conscious* ⇒ *knowing at least one truth*: If I state several obviously false statements (as evaluated by an adult conscious human) like that I have seven fingers on my hand or that I pinched you but said that I patted you instead and so on, you know that I am not telling the truth. Similarly, I cannot lie to you about the basic existence of objects, shapes, sounds and others. In other words, you already know several truths whenever you are conscious. We can also convince ourselves that the contrapositive is true, namely, that if you do not know even a single truth, you are not conscious.

   *Knowing at least one truth* ⇒ *minimally conscious*: Here, I will show the contrapositive, namely, that if you are not conscious, you do not know even a single truth, by considering all known cases when we are not conscious or partially conscious. Seven examples of these are mentioned in Section 1 (deep sleep, under general anesthesia, knocked unconscious, coma, severely drunk, sleepwalking and a newborn baby). In these cases, you do not *know* the existence of objects around you, even though you may avoid bumping into them when walking. As an example, under general anesthesia like, say, during a surgery, we do not know any truths including our body and our feelings. Only afterwards, but not during the surgery, it is sometimes possible to remember parts of the events which were memorized mechanically. Your ability to know basic logical facts is lost. In fact, you do not even know your own emotions, feelings, hands, legs and other parts of your body. When I pinch you, you will not know unless you wake up i.e., become conscious. The analysis is similar with the other six cases.

   In partially conscious states like when the anesthesia is wearing off, when you are about to fall asleep or when you are drunk, you are switching between a conscious and an unconscious state quite randomly. Therefore, your ability to know a single truth also switches accordingly.         Q.E.D

   The advantage with Claim 1 is that we can now study the theory of consciousness as a theory of truths and their relationships – a possibility, we never had before. This is the first result presenting necessary and sufficient conditions for consciousness, unlike any other theory. The rest of the paper explores the equivalent approach taking both an analytical and synthetical perspective.

## 3. Truths and fixed sets

The first step towards using the equivalent conditions of Claim 1 is to define what truths are and explain how our brain represents them within its neuronal architecture. This is a critical step because it involves taking an abstract notion like truth, described in a natural language like English, and representing it by a dynamical notion, described in terms of physical and chemical reactions (a requirement stated in Section 1.4).

### 3.1. Truths as abstractions

To formalize the abstract notion of truths, notice first that there are two common types of truths for a conscious being. Firstly, the existence of objects (tables or chairs), the relative relationships between objects (like legs attached to a chair), mathematical and physical laws, causality, geometric properties (like edges, angles and sizes), sound and visual patterns are accepted by most conscious beings as facts. When there is a chair in front of you, its physical existence is a truth irrespective of what you choose to call it. You can touch the chair with your hand and confirm its existence. Secondly, emotions, feelings, pain and other subjective or personal beliefs are truths unique to a given conscious being, which others may or may not agree. Your pain is true whether others agree with you or not.

We also group truths into several layers. Some like the geometric ones (edges and angles) are common across most objects, while others like the relative relationships between objects (legs attached to a chair) are distinct. Together, they create an intricate structure and relationships between our everyday truths in an abstract sense. For example, the truth "Sun rises in the east" requires several other interconnected truths like horizontal position, straight line, perpendicularity, eastern direction, existence of Sun, upward motion and day-night cycles. A conscious being (like an infant) cannot accept this new truth unless he represents all of the dependent truths and their network of relationships correctly. The network of relationships between truths keeps changing over time as well. For example, all truths associated to Santa Claus are intact for a child, whereas for an adult, they are broken down as new experiences alter them. Therefore, the objective here is to map all of these truths and their interrelationships within our brain network using dynamical interconnections between neurons.

For this, let me first define truths in generic terms beyond the colloquial usage to include everyday objects, events and experiences as discussed with the above examples. The intuition I use comes from the commonly accepted notion that *if an event keeps repeating the same way even under different variations, we tend to accept it as a truth (or reality)*. Conversely, if the event 'changes' under approximately repetitive experiences, we do not accept it as a truth. Therefore, repetitions and variations are both necessary to define truth. Excluding either one would alter the very meaning of truth significantly.

**Definition 1:** A truth is any representation like an object, event or a statement that does not change significantly (a) over a finite period and (b) under a sufficiently large set of dynamical transformations.

In this definition, the dynamical transformations capture the notion of variations while a large set of variations over a finite period captures the repetitions. It is clear that mathematical abstractions like the concept of numbers, your emotions, feelings, pain, space and time does obey this definition. A ghostly image does not satisfy this definition because it fails to repeat under several variations. Most lies also do not satisfy Definition 1 because variational repetitions produce significantly different results each time. When working with physical systems, evaluation of truth obeying Definition 1 is easier because experiments can be performed repeatedly and under variations. With abstract mathematical and computational systems, repeated computations and several collections of examples can be used to evaluate the truth of a statement.

From Definition 1, truths are allowed to change over long periods of time (like Santa Claus for a child or Sun going around the earth). The dynamical transformations in the definition correspond to the common variations we observe whenever an event containing the truth repeats. For example, let us pick one of the events as the average scenario ($\bar{E}$). Then, all other variations are expressed as $\{\Delta E\}$ around the average. Two common classes of transformations are space-time transformations and adaptive changes within our brain. Space-time transformations result from events like motion, turning and walking. For example, if I am looking at a table, all variations within my brain, triggered through my retina, for dynamical transformations like moving towards or away and turning my head as I continue to see a same table does not alter my notion of the table, i.e., it is a truth. The second sets of transformations are those that originate within our brain due to the adaptive changes in the dynamical network of neurons through new memories and experiences. For example, the concept of a table and the visual representation of it within my brain remains the same even after 10 years of new experiences.

## 3.2. Fixed sets – dynamical representation of truths

The next task is to identify the dynamical substructures within our brain that satisfy the above definition of truth. Our brain is a complex system with billions of subsystems interacting via neural firing patterns. From Definition 1, we are looking for a property of dynamical systems that (a) does not change for a while and (b) remains fixed even when the system is subject to several continuous transformations. One example is a stable fixed point. If we set the initial conditions to be near this fixed-point state, it continues to stay close enough for a while and even under minor variations because of stability – i.e., it satisfies Definition 1.

Brouwer's fixed-point theorem proves the existence of a fixed point (Granas & Dugundji 2003) for any static continuous transformation of a disk onto itself. However, when we consider dynamical systems (realistic ones, not mathematical idealizations), a stable fixed 'point' is not as common. Instead, a 'set' like a stable limit cycle, loops or invariant sets are more likely (Khalil 2001). Even though no single point in the limit cycle remains fixed, the entire set of points within the limit cycle does remain fixed (Fig. 2). If we setup the initial conditions in a neighborhood of the limit cycle, it will continue to stay within this set for a finite time and under several minor variations arising from repetitive experiments (because of stability) – i.e., it satisfies both conditions of Definition 1. I call these stable dynamical loops as *fixed sets*, generalizing the notion of fixed points.

From a synthesis perspective, we can create fixed sets using a graph-like network (Fig. 2) analogous to our brain network either in a computer or with mechanical components. In this case, the synthetic dynamical network can generate limit cycles, shown as $L_1$ and $L_2$ for an input falling on the nodes $I_1$-$I_3$ (Fig. 2). The specifics of the implementation of this synthetic system, though not described here, is easy to imagine with flow-based networks, network of computers, cellular automata and physical stable parallel looped (SPL) systems (Appendix A). From a graph-theoretic point of view, constructing a dynamical looped network as shown in Fig. 2 is direct and will be used frequently in this paper.

In the case of our brain, the formation of stable dynamical neural loops is guaranteed for most external inputs, as confirmed experimentally (Thompson & Swanson 2010). Loops are one of the most important, nontrivial and fundamentally pervasive dynamical structures of our brain. Nonlooped dynamics, on the other hand, are ephemeral. I exclude them here because they do not satisfy the conditions of being fixed under repeatable and variable scenarios, as required by Definition 1 for truth, even though they are necessary for linking firing activity between distant regions of our brain. Stable dynamical loops are the only pervasive structures that do satisfy all conditions of Definition 1.

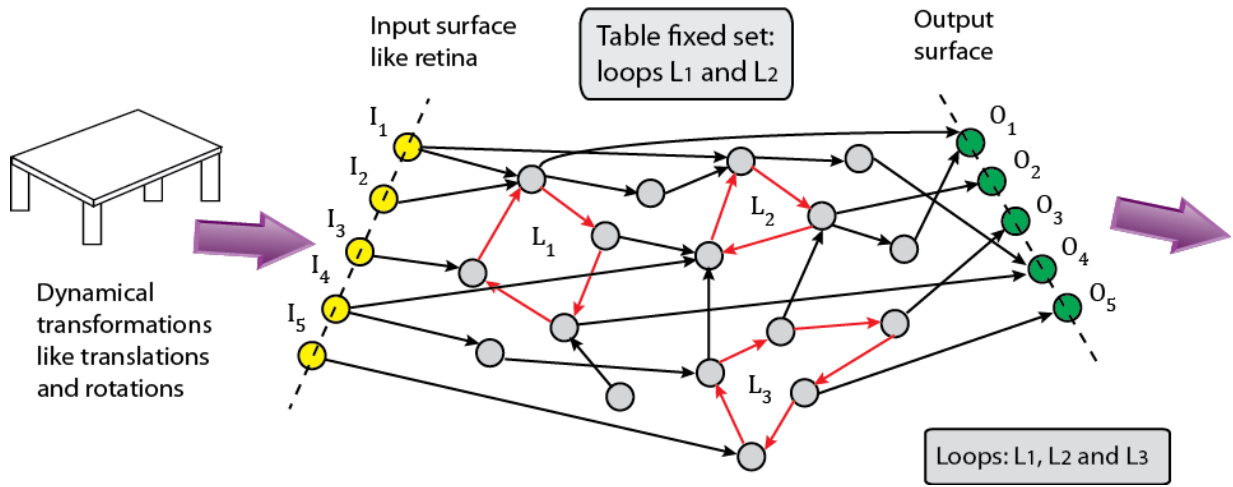

**Figure 2 | Sample stable parallel looped network.** External inputs like light from different objects fall on an input surface like the retina. Neurons shown as nodes fire in response and transmit the firing pattern deeper within the network. The set of dynamical loops triggered are $L_1$ and $L_2$ if the input falls on $I_1$-$I_3$ nodes. As the object is subject to minor dynamical transformations like translations and rotations, loops $L_1$ and $L_2$ are triggered as the common intersection. These two loops can then be identified as the fixed set for the table object.

The new theory on consciousness and truths is built on these basic looped dynamical components. The generalization is termed as stable parallel looped (SPL) systems. The construction of an infinite family of physical SPL systems is discussed briefly in Appendix A.

If you look at a table (or if a synthetic system processes the inputs from the table), the external inputs that fall on your retina trigger neural firing patterns that propagate within your brain network to form several dynamical loops. This is analogous to the triggering of loops $L_1$ and $L_2$ for inputs on $I_1$-$I_3$ shown in Fig. 2. Even when you close your eyes (or cutoff inputs in Fig. 2), some of these loops continue to fire compared to nonlooped firing patterns. For each variation around the average scenario of looking at the table (like moving closer, farther, left, right or turning your head), several thousands of the same retinal neurons, and, hence, a common set of dynamical loops continue to fire. Therefore, the *intersection* of the set of loops under each scenario, if nonempty, satisfies the Definition 1 of truth. I call this as the fixed set for the table.

We now uniquely identify the table with its fixed set. Our brain is capable of knowing (discussed later in Claim 5) about the existence of the table *if and only if* this fixed set is triggered. Indeed, if the table is distorted considerably or has a never-seen-before design, the above fixed set would not be triggered. In this case, the neural patterns generating from the retina is significantly different from all of the variations you have seen when you created your table fixed set. As a result, you do not recognize this object as a table. For example, a chair or a bookshelf is indeed significantly different from a table in terms of the neural firing patterns generated from the retina. As a result, you do not identify them as a table. Only variationally similar objects to table will be identified to a table. Each dissimilar object has a different fixed set. Another example of fixed sets is for different forms of handwriting, say, for the letter 'A'. All similar A's correspond to the same fixed set. Fixed set for 'A' is different from fixed set for 'B'. In a synthetic system like the SPL network of Fig. 2, it is quite easy to satisfy both the similarity and dissimilarity conditions above by altering the network suitably (see Figs. 3-4).

In effect, Definition 1, if satisfied, ensures that (a) similar objects generate nonempty intersection of loops and (b) dissimilar objects produce empty intersection (deeper in the brain). Let me state this connection between truths and fixed sets formally.

**Claim 2:** Fixed sets translate higher-level abstractions like truths into lower-level dynamics and vice versa within a stable parallel looped system.

## 3.3. Creation of fixed sets

It is well known that an infant has a sparse network at birth. His brain is still in the process of building structural neural connections between looped pathways. Therefore, for most objects and features, the intersection of loops generated from several simple spacetime transformations tends to become empty. However, with repetitive experiences over several months, he grows new neural connections using mechanisms already identified like growth cones that sense guidance molecules, the extension of dendrites and the creation of synaptic junctions (Kandel et al. 2000; Gazzaniga et al. 2008). As his brain network becomes dense through daily experiences, the possibility of having several neural loops with nonempty intersection and, hence, fixed sets improves naturally.

This can be empirically tested through experiments. Already, long-term potentiation, synaptic plasticity and detailed chemical basis for learning provide preliminary experimental validation (Lynch 2004; Kandel et al. 2000). This situation with the growth of neural connections is true even for an adult when learning a new topic. With a synthetic system like in Figs. 2-4, the creation and growth of SPL network can be manipulated through a training data set of images and sounds, say.

In this manner, a child builds fixed sets based on visual, tactile, auditory and other sensory inputs in addition to motor fixed sets. This takes several months of repetitive experiences (cf. Lynch 2004). Examples are fixed sets for edges, angles, colors, tables, chairs, repeated events, sounds, words in a language, actions like turning eyes or head, lifting hands or legs and others. The typical variations that help create these fixed sets are different accents for sounds of words, different objects with same colors and angles, different handwritings, different sources of same tastes, odors and touch, different situations for lifting your hand and so on. In a subsequent paper, I will describe unique mechanisms for creating each of these sensory-control fixed sets.

While the above fixed sets are created directly from external inputs, I will use the term ***meta-fixed sets*** to identify fixed sets created from other fixed sets instead. For example, the fixed set for the word 'table' is a meta-fixed set because it is not directly associated to a given external table. The advantage with meta-fixed sets (like for the word 'table') is that if you encounter a never-seen-before table, the links from the legs, shape and other context-related fixed sets will eventually trigger the same 'meta-fixed set' allowing us to identify the new table as a table. Meta-fixed sets are the most common types of truths we use with our natural language and other abstract representations.

The dynamical transformations necessary to create meta-fixed sets are, unfortunately, less common than for a fixed set because of fewer repetitive scenarios. For example, to create 'number' meta-fixed sets like '5', a child would need to encounter different number of objects under different scenarios repetitively for a long while. Similarly, most meta-fixed sets for language (like words) or mathematical abstractions (like numbers and measurements) take years to form. The repetitive variations are guided through teaching.

## 4. Self-sustaining membrane of meta-fixed sets

As truths are represented dynamically as fixed sets, the network begins to grow to form detailed interconnections. These interconnections are not random. Rather, they align with our experiences. This is true both in our brain and if we were to build a synthetic system like Fig. 2. In this section, I will explore this interconnected dynamical structure of fixed sets. Later in Section 6, I will show how to relate it to the problem of consciousness.

The importance of dynamical loops (not static structural loops of neurons like A-B-C-A, which always exist for any graph) and fixed sets are that they are the simplest, nontrivial and pervasive patterns. They already exhibit self-sustaining neural firing patterns (for, say, A-B-C-A) for a short while, even when the external inputs are turned off. How can we increase the total amount of self-sustaining time for a collection of loops? To see why this is important for consciousness, consider the time it takes to complete a single dynamical loop. If it is fast, our brain would be able to trigger many parallel as well as serial cascading sequences of fixed sets in one second. This gives us an ability to know several truths simultaneously at any given instant. This is required to provide a cohesive view of the external world. This in turn is necessary for consciousness. Therefore, the higher the number of self-sustaining active fixed sets, the greater our chance to become conscious and vice versa.

A direct way to estimate the total number of simultaneously excited fixed sets is by using brain waves. Under normal conditions, the electrical oscillations recorded by electroencephalogram (EEG) correspond to the sustained loops of firing activity in a given region. Nonlooped dynamics are ephemeral and appear as noise. Therefore, higher brain wave frequency corresponds to faster repeatability of looped dynamics. This is possible when we have a high number of fixed sets excited at a given instant and vice versa (empirical study is a future area of work). Gamma brain waves ($> 30$ Hz) correspond to a conscious state (Buzsáki 2006). As our brain wave frequencies become lower, i.e., from beta 12-30 Hz to alpha 8-12 Hz, theta 4-7 Hz and delta $< 4$ Hz, our conscious state lowers from alert to relaxed, drowsy and deep sleep, respectively. This corresponds to a decrease in the number of fixed sets. Indeed, when we approach an unconscious deep sleep state, we do notice that we know less number of truths (or, equivalently, fixed sets).

Therefore, to maintain gamma waves, how does our brain create and excite a large number of self-sustaining fixed sets at every instant? The following claim specifies one way to do this.

**Claim 3:** Consider an active system containing two loops. The dynamics of the entire system will self-sustain longer if the loops are positively linked together than if they were isolated (assuming no help from external inputs).

*Proof*: To see this, note first that each neuron and, hence, the loops of neurons have active energy source (as ATP in cells). This energy source is used to regenerate chemicals like the neurotransmitters and to drive the ions through the corresponding Na-K and $Ca^{2+}$ ion pumps in order to self-sustain the dynamics. Secondly, in the positively linked case (i.e., neural synapses linked through excitatory neurotransmitters instead of inhibitory ones), the outputs from the first loop provide inputs to the second loop and vice versa. Therefore, the additional neurotransmitter inputs from the second loop open new ion-channels on the first loop using the *active* chemical mechanisms (Kandel et al. 2000). This increases the total self-sustaining time for the linked-loops case. The result generalizes to all active looped systems.    Q.E.D

The above claim is also valid for any synthetic SPL system. If we take hundreds of loops, then from the above claim the total self-sustaining time for the entire system will be much higher if they are interconnected than if they are disjoint or sparsely connected. Furthermore, the stability of the entire interconnected looped system is guaranteed because of the directional nature of the input-output dynamics within the loops (see Appendix A).

There are two qualitatively different ways to link fixed sets. The first one can be seen as extending the existing set of loops *longitudinally* for a given abstraction (Fig. 3), while the second one appears to link two or more abstractions *laterally* (Fig. 4).

The former link helps a given fixed set grow and spread out deeper into the brain (or in a synthetic SPL system). This is necessary if we want to keep increasing the size of the collection of distinct objects we want to distinguish. The cellular mechanisms for these links are based on extensions to growth cones at the end of a developing axon known as filopodia and lamellipodia using guidance molecules like netrin, slit, semaphorins and ephrins (Kandel et al. 2000).

The latter link helps us associate two different events based on the simultaneity and repetitiveness of occurrences. The cellular mechanism identified here is long-term potentiation (LTP) (Kandel et al. 2000; Graham 1990). This mechanism has been shown to associate two or more distinct memories in the hippocampus, cerebral cortex, cerebellum and others (see associative learning or Hebb's postulate (Kandel et al. 2000; Gazzaniga et al. 2008; Graham 1990) summarized as 'cells that fire together, wire together').

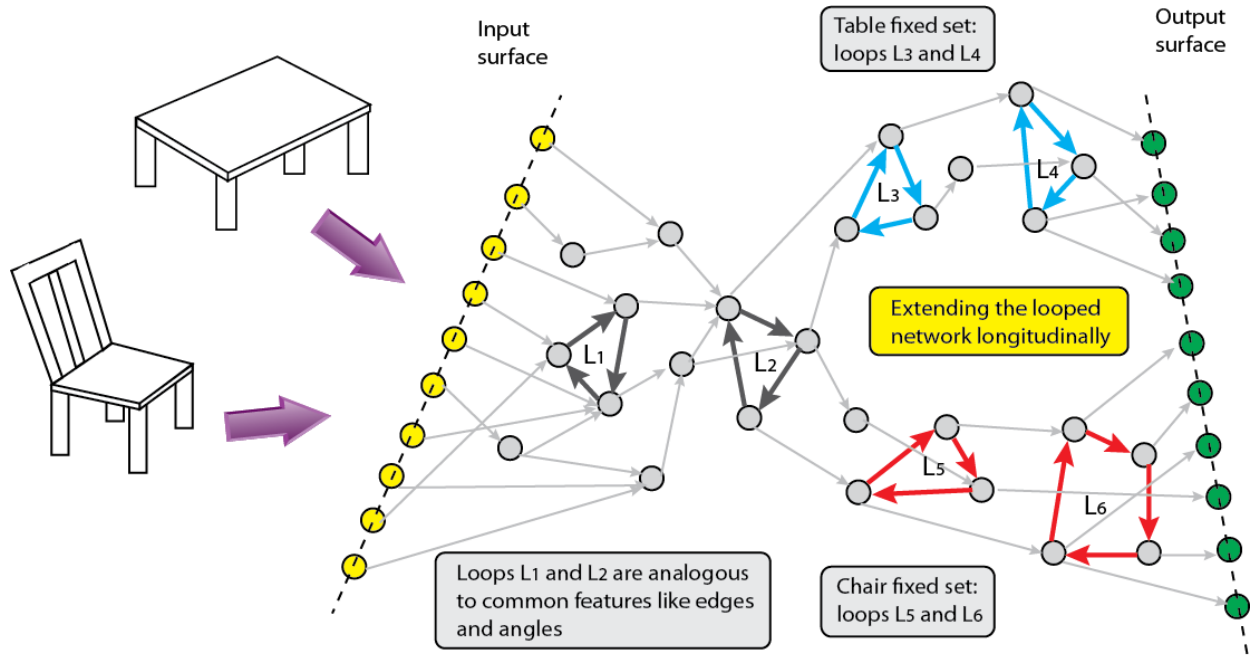

**Figure 3 | Evolution of stable parallel looped network – extension.** Over time, the set of loops extend and diverge into different directions longitudinally. This allows the network to distinguish several distinct objects easily. For example, consider loops $L_3$ and $L_4$ to represent a table. It can now be distinguished from a chair, represented by loops $L_5$ and $L_6$. The assumption is that the inputs for the table and the chair are different owing to the difference in shape so they trigger the above set of loops. Loops $L_1$ and $L_2$ are common for both and could not be used to distinguish them. Loops $L_1$ and $L_2$ are analogous to edges, angles and other basic features common to all objects.

During an infant age, the fixed sets are isolated and sparse. As an infant interacts with the external world, he maps the repeated experiences by creating several fixed sets. The above two linking mechanisms are now combined to create a dense 'mesh' of fixed sets, both longitudinally (Fig. 3) and laterally (Fig. 4). The links between loops are not arbitrary. For example, he links object-fixed-sets with shapes, boundaries, colors, textures and pattern fixed sets. Boundary fixed sets are linked to edge and angle fixed sets. Sound fixed sets are linked to image and word fixed sets (of a language). The motion fixed sets for your hands and legs are linked based on the common repeatable actions and so on. As an example, if a table fixed set is triggered, it causes a chair fixed set to be triggered automatically using these links and vice versa. In general, the dynamical relationships between fixed sets are directly related to the chain of events occurring at that instant. With a synthetic system like Figs. 3-4, we can easily control the creation of such specific links.

In addition to the above cellular mechanisms for creating links, we need new sensory-control mechanisms for vision, hearing, touch and motor actions to allow repetitions of events. These ensure that the same set of neurons are triggered repeatedly causing the chemical reactions and mechanisms to continue to occur until the desired connection is formed. Without repeatability of the events themselves, there is no easy way to access and trigger the firing of the same set of neurons hiding deep within the brain. I will discuss several of these sensory-control mechanisms in subsequent papers.

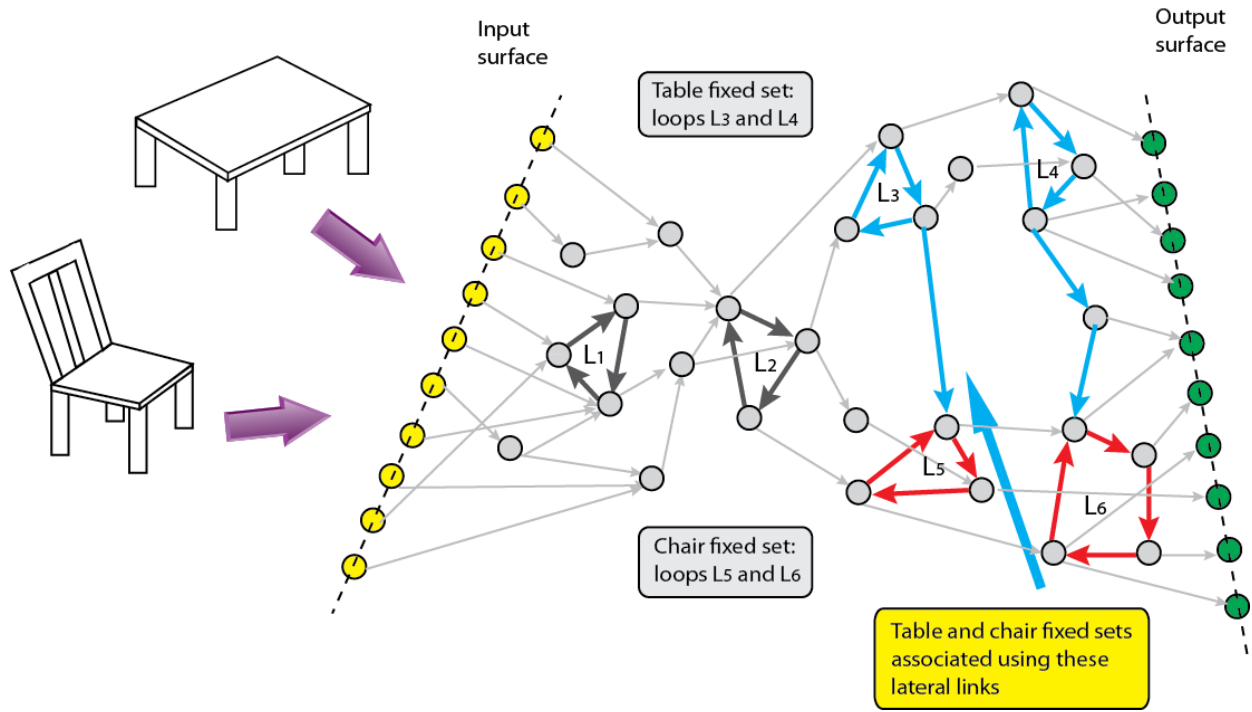

**Figure 4 | Evolution of stable parallel looped network – association.** Over time, with repetitive occurrence of events, the association is memorized within the network using lateral links. For example, as tables and chairs are always together, loops $L_3$ and $L_4$ (table fixed set – Fig. 3) are now connected laterally to loops $L_5$ and $L_6$ (chair fixed set). As before (Fig. 3), loops $L_1$ and $L_2$ are analogous to edges, angles and other basic features common to all objects.

Using these mechanisms, an infant builds a large repository of fixed sets and links between them during the first few years of his life. With multiple intersecting pathways through several experiences, it becomes possible for fixed sets themselves to form larger loops (i.e., loops of loops). A simple thought, a quick glance or a small action will trigger initially a few fixed sets within this collection. However, as they propagate via dense interconnections, our brain produces an 'explosion' of excitations. Generalizing Claim 3, we see that a well-connected network of loops self-sustains for a long time in which the neighboring loops help each other to continue the dynamics (Figs. 3-4).

I call this unique and well-defined dynamical state as the self-sustaining membrane of meta-fixed sets. The membrane receives sensory information (cf. input surface – Figs. 2-4) causing it to trigger control actions (cf. output surface – Figs. 2-4). This in turn supplies new sensory information triggering additional control actions, thereby producing a stable self-sustaining looped process. During this stable dynamical state, the membrane shrinks in some regions and expands to other regions depending on the external signals or the internal inputs (thoughts) at that instant. The static network of fixed sets should have a threshold density in order to have a self-sustaining dynamical membrane. Let me state the existence of this unique state formally as the following claim. The mechanisms for the construction of this state will be considered in a subsequent paper.

**Claim 4:** There always exists an active stable parallel looped dynamical system that self-sustains as a whole, in which the sensory information triggers control actions and the control actions triggers what it senses, continuously in a stable looped manner for a sufficiently long time. This dynamical state is termed as the self-sustaining threshold membrane of fixed sets.

One way to characterize the threshold state of the membrane is in terms of brain waves. For a threshold state, the fixed sets of self-sustaining electrical activity must produce gamma brain waves ($> 30$ Hz), say, by exciting enough fixed sets per second. If the number of fixed sets drops low enough to only result in, say, delta brain waves of less than 4 Hz, you will no longer have a self-sustaining membrane. This corresponds to a deep sleep state. During REM sleep, the brain wave pattern increases back to gamma range through internal self-excitations, thereby reconstructing the self-sustaining threshold membrane. Such a unique dynamical state is exhibited by a class of stable parallel looped dynamical systems, which includes humans and other animal species (like mammals). Recent experimental study did measure such a self-sustaining electrical activity in our brain (Fröhlich & McCormick 2010).

Let me briefly evaluate how the membrane behaves when you are less conscious. For example, an infant does not have a dense network of fixed sets. His membrane is not well connected to self-sustain even if external inputs reach the sensory surfaces regularly. His threshold membrane pops in and out of existence. When you are about to fall asleep, the number of fixed sets triggered reduces as you close your eyes and you stop processing, say, less audible sounds in the room. Your membrane shrinks and does not self-sustain. When you are under general anesthesia or severely drunk, specific chemicals block the ion-channels of neurons. As a result, the looped dynamics is less likely to self-sustain or produce cascading effects. The membrane shrinks in this case as well. If you are in coma or you are knocked unconscious, critical junctions are either damaged or their blood supply cut-off. It is now difficult to self-sustain the looped dynamics across wide regions of our brain that pass through these critical junctions. In each of these cases, the membrane is either weak or tears down.

## 5. Abstract continuity

The above discussion on truths and fixed sets resulted in the creation of a self-sustaining threshold membrane of meta-fixed sets. The next topic needed for Claim 1 is to address how we attain a feeling of knowing (Fig. 1). For this, we first require a new notion of continuity applicable to abstractions, which I will discuss now. In the next section, I will combine both the dynamical membrane and the abstract continuity to provide necessary and sufficient conditions for a feeling of knowing (Claim 5).

Recall that abstractions are represented dynamically as fixed sets (Section 3). Therefore, if we pick a given observable event, typically, expressed in abstract terms, we can use this one-to-one mapping to produce a corresponding chain of dynamical fixed sets. Now, since the dynamics already has a well-defined notion of continuity, it follows that continuity for abstractions is well-defined as well. However, this is a weak form, not sufficient for feeling of knowing. I will now discuss a stronger notion called abstract continuity.

Consider sentences in a language like 'I like to eat pizza tonight' versus 'pizza to like I tonight eat'. We understand the former sentence but not the latter, even though the latter sentence is a simple rearrangement of words from the former sentence. We understand each individual word in the latter sentence and, yet, do not understand the entire sentence. Clearly, the former sentence is grammatically correct unlike the latter. However, why should conscious humans (even a 2-3 year old) ever understand grammatically correct sentences? In general, it is easy to create grammatically correct sentences that you have never heard of (like sentences in this paper) and, yet, you understand them. Why or how? The situation is true with partial sentences if they are arranged in an approximately grammatically correct way (like say, 'want pizza tonight').

As another example, consider a jigsaw puzzle. The incorrect arrangement of the pieces, analogous to the incorrect arrangement of words in the above example, makes us not understand the picture. As we start arranging them in the correct order, we begin to attain a feeling of knowing of the entire picture (analogous to grammatically correct arrangement). Similarly, consider an event in which we are trying to recollect a person whom we just forget, but know that we have seen him before. We can say analogously that the thoughts and events are incorrectly arranged that we are not able to attain a feeling of knowing of that person. As we start to arrange them in a 'correct' order, we suddenly attain the feeling of knowing. Our objective is to understand what the necessary and sufficient conditions are for all such cases – correct arrangement in sentences, correct arrangement of jigsaw-like images, correct arrangement of sound variations, correct arrangement of events and so on – to attain a feeling of knowing (Claim 5).

The answer common to all of these cases depends on a definition of continuity within abstractions. I will define it now using sentences as the primary motivating example. The generalization to all other cases of feeling of knowing will be discussed later in Section 6. The key idea in defining abstract continuity is in linking abstract sentences to our ability to *predict at every instant* (a dynamical notion).

In the context of stable parallel looped systems, I define prediction in the following way. If the system receives sensory inputs corresponding to an external event, I say that it predicts the future states of the external event if the system triggers the corresponding future fixed sets in approximately the same time order. For example, when a ball is bouncing around, an adult predicts its future states by triggering corresponding future fixed sets of the motion of the ball correct to, say, about a second into the future, ***at every instant*** (not just the final location or trajectory). If the event is already continuous and dynamical, prediction at every instant is natural. My goal is to show that the same continuity applies for discrete words of a sentence in a language as well.

In the sentence 'I like to eat pizza tonight', when I hear the word 'I', I am able to predict, 'I want…', 'I like…', 'I am…' and others as immediate possibilities. Next, when I hear 'I like', I can predict 'I like music…', 'I like to…' and others as possibilities. For 'I like to', the nonempty choices are 'I like to do…', 'I like to play…' and others. For 'I like to eat', the nonempty predictions are 'I like to eat candies…', 'I like to eat ice cream…' and so on. Similarly, we have nonempty predictions for the rest of the sentence as well. The choices need not be correct at every instant relative to the sentence you are hearing. Just being nonempty is sufficient.

On the other hand, consider the sentence 'pizza to like I tonight eat'. When I hear 'pizza', the nonempty predictions are 'pizza is…', 'pizza has…' and others. However, the moment I hear the next word 'pizza to', I have lost all my predictions. We have an empty set of predictions. The situation does not improve when I hear 'pizza to like' or 'pizza to like I' and so on.

This is the fundamental difference between any grammatically correct sentence and an ill-formed sentence for an adult conscious being. All sentences we do understand have nonempty predictions if we break them word-for-word in a linear order like above. This condition is, however, not sufficient for understanding a sentence (Claim 5). For now, I will use the above example to define abstract continuity as follows.

**Definition 2:** Abstractions are said to be continuous if their dynamical representations produce nonempty, and possibly incorrect, predictive meta-fixed sets along most linear subsequences. These subsequences are either implicitly present (like with linear sentences in a language) or explicitly created (like with linear paths when scanning a 2D image with our eyes) within the abstraction.

Notice that this definition generalizes to jigsaw-like images mentioned earlier and, in fact, to all abstractions. Predictive ability is a universal notion applicable to all events and experiences. What is unique is to show how predictions play a direct role for the problems of understanding and knowing of any event (Claim 5).

## 5.1. Verifying abstract continuity

To check abstract continuity for a discrete abstract sentence, we first translate it into a dynamical representation of linked fixed sets within a given stable parallel looped system. Using standard 'dynamical' continuity of neural firing patterns, we verify if there are nonempty predictive meta-fixed sets along the entire pathway at *every* fixed set and most subsequences locally, as we did with the examples considered previously. I say that abstract continuity is broken if and only if the above dynamical continuity is broken. As mentioned in the definition, it is important to have nonempty predictions for 2-word (like 'to eat' and 'eat pizza' as opposed to 'like I' and 'I tonight'), 3-word (like 'like to eat' and 'to eat pizza' as opposed to 'like I tonight') and other subsequences as well, not just for 1-word sequences considered in the previous discussion.

## 5.2. Creating abstract continuity

How does our brain create abstract continuity for a given event in the first place? If you are learning to catch a ball through repeated scenarios, we identify an average scenario and several variations around it. For example, the ball reaching you correctly is an average scenario while other mistakes like reaching too close or too far from you are common variations. Through each of these experiences, our brain stores nonempty predictive meta-fixed sets for the ball-catching event (cf. synthetic systems like Figs. 3-4). These predictions let us reach for the ball quite accurately even in a new situation. Similarly, when you are learning a new topic in abstract mathematics, you create fixed sets for each of the primitive concepts first (like groups, rings and fields). Through several examples (like real numbers and matrices), you then create *predictive* meta-fixed sets around each fixed set. Now, when you encounter a new theorem, each of these nonempty predictions will give rise to abstract continuity.

The situation is similar when a child learns a language. Initially, he builds abstract continuous paths for the objects and actions directly using vision and touch (along with the vision and touch-based fixed sets). He then links or associates the visual and tactile abstract continuous paths with words in a language. Over months of experiences, he slowly transitions to abstract continuity within language. When formulating new sentences on his own, he does not need to create and follow grammatical rules. Instead, he can use the looped network created by the abstract continuous paths. Therefore, hearing sentences in a language that merely 'describes' an event is as real as 'observing' the physical event itself because of these abstract continuous paths.

Creating abstract continuity in each of these cases is a slow process. The abstract continuous pathway for a given event stored within our brain is not like a thin trajectory, but a 'wide band' around the average pathway. Generalizing from these examples, we now say that our brain creates abstract continuous paths for memorizing physical events, understanding mathematical statements, learning sheet music or a new language, designing engineering structures and other common everyday tasks using sensory-control mechanisms and repetitive experiences.

In the case of a synthetic system (like Figs. 2-4), creating abstract continuous paths is a manual process of linking fixed sets based on different variations of training examples. For sentences in a language, this involves linking different word fixed sets in such a way that the system has only those nonempty pathways at a given fixed set that correspond to grammatically correct sentences. This involves picking a training set of grammatically correct sentences, say, from the story books a child reads at a young age. For grammatically incorrect sentences, the nonexistence of predictive meta-fixed sets results in a discontinuity.

One interesting case to notice is that a child forms grammatically incorrect sentences sometimes even though he is only exposed to grammatically correct sentences. This is because the evolving cumulative network automatically has extraneous links between meta-fixed sets even though he has never heard these sentences. As a result, he believes that it is a meaningful sentence. This is usually corrected by a teacher or a parent. This teaching process *deletes* some of the existing links thereby pruning the network to stay as close to grammatical correctness as possible. This pruning step can be performed manually even in a synthetic system like in Figs. 2-4.

## 6. Feeling of knowing

We now come to the last topic needed for Claim 1, namely, how we attain a feeling of knowing for events. In the previous section, we have seen that abstract continuity captures the unique difference between a grammatically correct and an incorrect sentence. However, attaining a feeling of knowing requires a bit more. The claim below does not say 'why' we have a feeling of knowing similar to Claim 1 for minimal consciousness (or any other scientific fact). It only guarantees the existence of a feeling of knowing when specific conditions are satisfied and vice versa. The result is yet another example of a subjective notion that is represented equivalently as unique dynamical properties of a special subset of stable parallel looped systems. Emotions and other types of feelings are not covered by this claim (will be discussed in a subsequent paper).

**Claim 5:** Knowing an abstraction (say, an event, object or a sentence) is equivalent to the existence of a self-sustaining membrane of meta-fixed sets and the triggering of an abstract continuous path from the meta-fixed set representation of the abstraction to the dynamical membrane.

*Proof*: Consider looking at a picture when we are conscious versus when we are not (like when the general anesthesia is wearing off or when severely drunk). Most of the times, we instantaneously attain a feeling of knowing of the picture when we are conscious. However, there are times when we are distracted with other thoughts that we do not know what we were looking at. Only when we turn our attention back at what we were looking previously, the feeling of knowing is attained. To relate this example to Claim 5, I say that our, possibly distracted, conscious thoughts make up our initial dynamical membrane. The image input through our eyes trigger a collection of meta-fixed sets. If the intersection between these fixed sets and the dynamical membrane is empty, we do not attain a feeling of knowing. If, however, we turn our attention back and actively think about the picture, I claim that we are expanding the neural firing regions of the dynamical membrane to overlap and intersect the corresponding fixed sets initiated from the retinal images. If and only when this happens, I will show that we attain a feeling of knowing.

*Knowing ⇒ membrane + abstract continuity*: Let us assume that we are observing an event and it produces a feeling of knowing. From Claim 1, we are minimally conscious, as we know, at least, this current truth. However, the truths are interconnected to form a network within stable parallel looped systems. It is illogical to say that an infant 'knows' the event when he does not form a membrane yet. Minimal consciousness imposes structural and dynamical constraints within this network of truths as mentioned previously. At minimum, these constraints guarantee the existence of a self-sustaining threshold membrane of meta-fixed sets.

Now, to show that we need abstract continuity for the event, I will do so by contradiction. I will assume that we know the event, but do not have abstract continuity. I want to show that this is illogical. With no abstract continuity, there will be no predictive meta-fixed sets at every instant along some linear subsequence.

The first result is that you do not have any memory of the current event. The reason is that memory (or fixed set) is never isolated and it always creates abstract continuous paths. For example, a few related memories stored for any event are: (i) when you observed the event, (ii) other events occurring around you and (iii) your emotional state like happiness or sadness. All such related memories are linked together naturally. These will later give rise to nonempty predictions, violating our assumption. Note that a lack of memory, by itself, is not contradictory with our ability to know an event. For example, several sentences in this paper are new and, yet, you know or understand them.

Next, I will breakdown the original event into subcomponents (like words) along a linear sequence (like a sentence). This is always possible by choosing components along a natural timeline. If every subcomponent is new (i.e., without past memories), then knowing the entire event is illogical. Therefore, we have two possibilities: (a) some subcomponents are new with no past memories or (b) all subcomponents have past memories while the linear sequence itself is new. Case (b) is analogous to the ill-formed sentence of the previous section. Knowing such a sequence is illogical and, therefore, contradicts our assumption.

To eliminate case (a), note that we eliminate some of the new subcomponents if they are not critical. A 3-year old understands a sentence even if he does not know every word in the sentence. If all new subcomponents are of this type, we have a contradiction. To see this, when we ignore the new subcomponents, the event is either made of memorized subcomponents which already have abstract continuity or they reduce to case (b). Both situations lead to a contradiction. Therefore, we cannot ignore a few new subcomponents with no memories. We continue breaking them down into smaller subcomponents until there is no value in doing so. We now regroup them into categories such that we cannot know each of these categories as they contain at least one unknown critical subcomponent. This implies that we have a linear sequence of categories each of which is unknown while the entire sequence itself is known. This is a contradiction. As a result, case (a) is also not possible. This completes the proof in one direction.

*Membrane + abstract continuity* $\Rightarrow$ *Knowing*: A brief outline of the proof is as follows. First, I show this result for one family, namely, that of sentences in a language. Next, I pick two additional families, dynamical events and abstract ones. I show how to decompose events within these families into primitive components and linearly interconnected sequence of events analogous to words and sentences in a language. Using such a mapping, I show the result to be valid for these two families as well. Next, I generalize the decomposition approach to all other cases. When this happens, the primitive components will be shown to be part of the membrane while the linear sequence of events becomes part of the abstract continuous paths. Therefore, together they give rise to the feeling of knowing for all cases.

Let me now discuss these steps in detail. As seen in Section 5 with knowing sentences in a language, knowing the simplest words is equivalent to representing them as fixed sets and creating abstract continuous paths with real physical objects and actions. These 'primitive' words are recursively linked together to form a complex network, which becomes part of the membrane (i.e., you see it as a whole), while knowing the sentence made of these words requires abstract continuity.

Let us now generalize sentences in a language to other examples. In the previous section, we have already seen the similarity with other examples like jigsaw images, odd arrangements of objects like tables and with recollection of events. As the first family of systems, consider building physical dynamical systems. During our childhood, we learn to create abstract continuity for the basic operations that make up any given machine or experiment. These are, say, the set of dynamical operations like assembling parts, gluing, bending, stacking, breaking, moving and applying forces as well as familiarity with a set of materials like plastics, paper, wood, metals, bolts, capacitors, chemicals and cement. Therefore, using these primitives, we have a feeling of knowing for the construction of other machines involving similar components and operations (cf. do-it-yourself kits). In fact, even reading the setup of an experiment is sufficient to gain abstract continuity and a feeling of knowing without having to build it.

Next, in order to generalize to all experiments, we represent each dynamical experiment as components and processes. The components refer to real physical objects. We can see, hear and touch them. Their existence cannot be denied because we have multiple sense organs to detect them consistently and reliably. They are also reused within multiple designs. The processes are sequence of operations that you apply on the components. These processes obey the laws of nature. You apply the processes in a specific order either serially or in parallel to construct the final design. As with the components, we reuse these processes in other designs. We can now pick any design and represent how to create it in terms of a set of reusable components and processes.

These components and processes can be further divided several times until we have a collection of primitive subcomponents and subprocesses common across all objects and experiments. In some cases the primitive components and processes become obvious only when arranged in a haphazard way and then rearranged correctly later (cf. the jigsaw-like puzzles).

The second set of examples is with abstract events. For abstract mathematical statements, a concrete set of examples (like using integers, functions or matrices) take the equivalent role of experiments in dynamical systems. They help build the necessary abstract continuity for each new concept like groups, rings, fields or vector spaces. With abstractions like discussions, debates and arguments, the primitive sets are analogies for which we already have abstract continuity using relevant historical facts or cases (as doctors and psychiatrists use) and precedents (as lawyers use).

Now, each of these cases is analogous to the case of grammatically correct sentences if we map components to words, both of which are reused. Components that cannot physically exist in the real world are analogous to gibberish words, which we clearly eliminate. The processes obeying the laws of nature are equivalent to grammatical rules of a language. Since the desired result is already true for grammatically correct sentences, I say that the result remains true, via generalization, for both dynamical and abstract events. Next, as all events occurring around us every day can be classified as dynamical events, abstract events or a combination of both, the above generalization now applies to all events.

Now the last step is to see how to represent the primitive components and processes themselves. If we look at a language, all words are defined in terms of other words and a set of primitive words. The primitive words themselves are associated directly with dynamical objects and events that most conscious beings agree as self-evident. For example, a straight line is self-evident because we can draw it on a piece of paper to verify its definition. The sound for the letter 'A' (or any word) is self-evident because we can generate a sound and hear it back to confirm that it is the same sound. A shape of the letter 'A' (or any word) is self-evident because we can write the shape on a piece of paper and look at it to confirm again. Similarly, colors, edges, shapes, tastes, odors, words to describe emotions and definitions or axioms in mathematics are basic abstractions chosen to be self-evident for most conscious beings. They are expressed directly in terms of tangible objects and physical dynamics that our sensory organs can detect reliably, repeatedly and at will. The self-evident rules that combine the primitive abstractions are obvious in mathematics (like axiomatic principles) and science (like the laws of nature). These rules do change over time just as grammatical rules can change in a language.

What makes a primitive component or process so obvious that you simply take it for granted? The primitive components are like colors, edges, angles, sounds or the basic sensory inputs. The primitive processes are like lifting your hand, turning your eyes or head or the basic control actions. These are the most repeatable phenomena in our everyday life. Each of these primitive components is represented and stored in our brain as fixed sets. They are used to *define* new terms and concepts. When the fixed sets are networked together, they create abstract continuous paths. These paths become the primitive processes. The network of abstract continuous paths evolves to become a membrane of meta-fixed sets. Once our brain creates a threshold membrane, it can express every other event in terms of these primitive components and processes. The self-sustaining membrane works now as a whole entity (like a 'picture') instead of just a collection of seemingly disjoint individual subcomponents (like 'thousand words'). Together with abstract continuity, they are now sufficient to attain a feeling of knowing. This completes the proof.  Q.E.D

**Claim 6**: The self-sustaining membrane of meta-fixed sets is equivalent to a dynamical network of abstract continuous paths.

**Claim 7**: A system is minimally conscious if and only if it has a self-sustaining threshold membrane of meta-fixed sets.

**Claim 8**: Knowing a single truth is equivalent to knowing an entire membrane of truths simultaneously.

Claim 6 follows directly from the proof of Claim 5. Claim 7 follows from Claim 6 and Claim 1. Claim 8 follows from Claim 6 and Claim 7. We never know truths one after the other, in a discrete manner (like knowing just a tree or some 10 facts and nothing else). The interconnections between truths force us to know several truths, specifically, an entire membrane of truths, simultaneously or none at all.

## 7. Conclusions

Whenever we try to locate the source of consciousness, we always seem to face one fundamental issue. If someone or something in our brain lets us observe, perceive and make decisions to perform specific actions, who or what tells this someone or something to do these tasks? We can continue asking this question *ad infinitum*. What kind of an answer would eliminate this paradox? Claim 7 gives us a way out of this infinite regression. To see this, note that the main source of difficulty with the idea of someone or something as the source of consciousness is that we are trying to associate the notion with a *static localized or distributed structure* within our brain.

Claim 7 says that this is an incorrect view. Instead, it says that our brain network dynamics should evolve over time to reach a unique *dynamical state*. This dynamical system should have a special membranous structure that self-sustains for a long time. This membrane is connected. It should keep a minimal set of representations of facts like geometry (angles, edges and shapes), motion, our own body relationships, basic set of actions, space and time in an excited state. It extends and contracts from one region to the other depending on what your external sensory inputs are and what your internal thoughts are. If this dynamical membrane stays intact, you are conscious. If this membrane tears down, you necessarily become unconscious. The unique property of this membrane is that it behaves coherently sensing itself, controlling itself and in which the sensing affects the control and vice versa, as one big loop. This looped process is implemented using millions of smaller interconnected neural loops, loops of loops and so on.

When we now ask who is presenting a cohesive view in front of us, the answer is the membrane itself. It controls your eyes, head and your body and makes you turn in specific directions to get new information from the external world. Subsequently, the membrane itself senses all the information it receives after you have turned your eyes or head. The membrane now contracts or expands to other regions of your brain as it processes the new information. This causes the membrane to control your body once again to gather new information, which in turn senses new information, followed by expansion/contraction and so on in a repeating loop. This looped process ensures that the membrane keeps receiving inputs continuously, thereby maintaining a core network of neurons in an excited state. The looped process self-sustains its own existence while consuming cellular energy.

The self-sustaining looped nature of the dynamical membrane eliminates the problem mentioned above in which we have one layer leading to another layer *ad infinitum*. With the self-sustaining membrane view, all layers are the same, namely, the dynamical membrane itself. The membrane is a self-sustaining causal loop – the cause produces effects, which in turn produces new causes and so on. When such a giant causal loop includes all types of visual, tactile, sound, geometric patterns, your own body inputs, space and time, I say that resulting structured dynamical system is minimally conscious (Claim 7). If it tears down, you become unconscious.

Consciousness is a dynamical *state*, not a specific region in the brain. In this paper, I only focused on what consciousness *is*, not on the underlying mechanisms that let us become conscious or that let us create synthetically conscious systems. These specific mechanisms will be discussed in subsequent papers.

Three of the important features of the new theory are: (a) deriving necessary and sufficient conditions so we could study consciousness using an alternate equivalent approach while still offering guarantees towards a complete solution, (b) transitioning back-and-forth between abstractions and dynamics so we understand how abstractions emerge from dynamics and vice versa and (c) picking stable parallel looped dynamical systems as the correct class both for the analysis and synthesis of systems that have millions of interacting dynamical subcomponents.

Claim 7 implies that several other animal species satisfy the necessary and sufficient conditions using vision, touch and its own body fixed sets. The term minimality is, therefore, necessary to distinguish it from human consciousness. The specific abstractions stored within the dynamical membrane (like patterns based on language, light, sound, touch and others) is ultimately responsible for distinguishing human consciousness from the minimal notion. I will discuss them in subsequent papers.

## References

1. Baars, B. J. (1988) *A cognitive theory of consciousness*. Cambridge University Press.
2. Buzsáki, G. (2006) *Rhythms of the Brain*. Oxford University Press.
3. Chalmers, D. (1996) *The Conscious Mind: In Search of a Fundamental Theory.* Oxford University Press.
4. Chomsky, N. (1957) *Syntactic structures*. de Gruyter Mouton.
5. Crick, F. & Koch, C. (2003) A framework for consciousness. *Nature Neuroscience* 6: 119-126.
6. Dennett. D. (1992) *Consciousness Explained.* Back Bay Books.
7. Edelman, G. M. (1987) *Neural Darwinism: The theory of neuronal group selection*. Basic Books.
8. Fröhlich, F. & McCormick, D. A. (2010) Endogenous electric fields may guide neocortical network activity. *Neuron* 67:129-143.

9. Gazzaniga, M. S., Ivry, R. B. & Mangun, G. R. (2008) *Cognitive Neuroscience: The Biology of the Mind*. W. W. Norton & Company.
10. Graham, R. B. (1990) *Physiological psychology.* Wadsworth Publishing Company.
11. Granas, A. & Dugundji, J. (2003) *Fixed Point Theory*. Springer.
12. Hameroff, S. (2006) Consciousness, neurobiology and quantum mechanics. In: *The emerging physics of consciousness,* (Ed.) Tuszynski, J. 193-253, Springer-Verlag.
13. Jaynes, J. (2000) *The origin of consciousness in the breakdown of the bicameral mind.* Mariner Books.
14. Kandel, E. R., Schwartz, J. H. & Jessell, T. M. (2000) *Principles of Neural Science*. McGraw-Hill Companies.
15. Khalil, H. K. (2001) *Nonlinear Systems.* Prentice Hall.
16. Koch, C. (2004) *The Quest for Consciousness: A Neurobiological Approach.* Roberts and Company Publishers.
17. Koch, C. & Tononi, G. (2008) Can machines be conscious? *IEEE Spectrum* 45: 55-59.
18. Lynch, M. (2004) Long-term potentiation and memory. *Physiological Reviews* 84 (1): 87-136.
19. O'Regan, J. K. & Noë, A. (2001) A sensorimotor account of vision and visual consciousness. *Behavioral and Brain Sciences* 24: 939-1031.
20. Ravuri, M (2011) Evolution of Order – An Application of Stable Parallel Looped Networks *arXiv:1102.1407v2*.
21. Roorda, A. & Williams, D. R. (1999) The arrangement of the three cone classes in the living human eye. *Nature* 397:520-522.
22. Thompson, R. H. & Swanson, L. W. (2010) Hypothesis driven structural connectivity analysis supports network over hierarchical model of brain architecture. *Proceedings of the National Academy of Sciences, USA.* 107(34):15235-15239.
23. Tononi, G. (2004) An information integration theory of consciousness. *BMC Neuroscience* 5: 42. http://www.biomedcentral.com/1471-2202/5/42

# Appendix A – Stable parallel looped systems

The class of stable parallel looped (SPL) systems is a special family of dynamical systems with a unique property that they can be created to remain stable even if we intentionally increase the complexity to any specified level. When studying complex systems (i.e., those that have millions of interacting dynamical subcomponents with a high degree of order) like cells or our brain, the first prerequisite is to know how to construct such complexity while ensuring stability. All man-made dynamical systems tend to collapse quite easily when the complexity becomes this large. SPL systems are the only known class of dynamical systems that avoid such collapse. In this appendix, I will describe what SPL systems are; specify an infinite family of systems with arbitrary complexity and show how to create them synthetically.

**Definition A1:** A dynamical system is said to be a stable parallel looped (SPL) system if and only if the following conditions hold: (a) *existence of dynamical loops* – there exists parts of the physical or chemical dynamics of the system that recur approximately the same with a time-varying period $T(t)$ for a sufficiently long time and, if necessary, by supplying energy and/or material inputs repeatedly, (b) *parallel interacting loops* – the system has one or more loops that exchange inputs and energy with one other and (c) *stability* – there exists sufficiently small disturbances for which the system does not collapse (like, say, bounded inputs producing bounded outputs).

In order to exclude systems that die down quickly, I will typically impose the following condition as well – an SPL system should be stable sufficiently longer than the time period of the slowest looped dynamics (say, at least 10 times longer). Furthermore, I only require approximate repeatability and periodicity with no necessity to continue the dynamics indefinitely.

The simplest example of an SPL system is a swing in a playground. A person pushes at regular intervals to supply small energy to continue the oscillatory (or looped) motion. The other two conditions in the definition are trivially satisfied. Other examples are limit cycles, periodic, cyclic and oscillatory ones (like pendulums and waves). With chemical systems, examples of SPL systems are chemical oscillatory reactions like Belousov-Zhabotinksy reaction, Briggs-Rauscher reaction and interconnected cycles like Kreb's cycle, Calvin-Benson cycle and the more general metabolic networks within living beings.

## A1.   Physical SPL systems

Fig. A1(a) show a simple SPL system with balls rolling down an inclined plane and a conscious human closing the dynamical loop. To eliminate the involvement of a human and improve the creational difficulty partially, Fig. A1(b) shows a modified design that curls the inclined plane to form a structural loop. If the energy lost through friction is small, random energy bursts can occasionally cause the balls to reach the top and continue the dynamics for a few loops.

One problem with Fig. A1(b) is that the balls move in both forward and reverse direction. A single stable point at the bottom is the cause of this attraction from both directions. The resulting collisions hinder the overall motion with all balls settling at the bottom (the lowest-energy state). To restart the looped dynamics once again, every single ball needs a sufficiently large amount of energy. Fig. A1(c) is a redesign with several hills and valleys along the looped path (cf. roller coaster designs). With multiple stable points (valleys), the balls store part of the external energy as potential energy at these intermediate heights unlike Fig. A1(b). Restarting the dynamics becomes easier. The balls also have less average number of collisions. Therefore, this SPL system exhibits dynamics longer than the previous design since the occasional random energy from the environment is more effectively trapped.

### a. Simple SPL system with help from a human

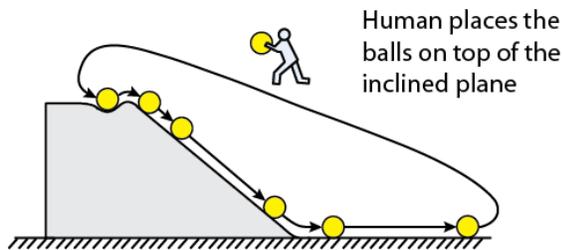

### b. SPL system with no help from human

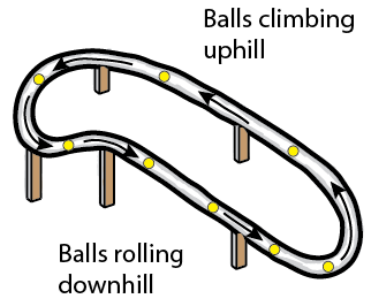

### c. SPL system with a single valley changed to multiple stable valleys

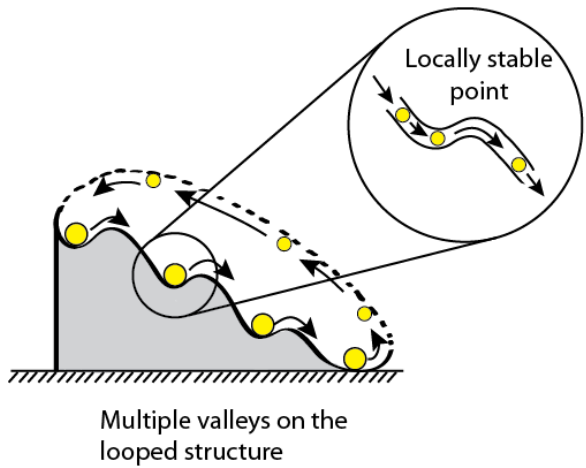

### d. SPL system with valleys changed to loops

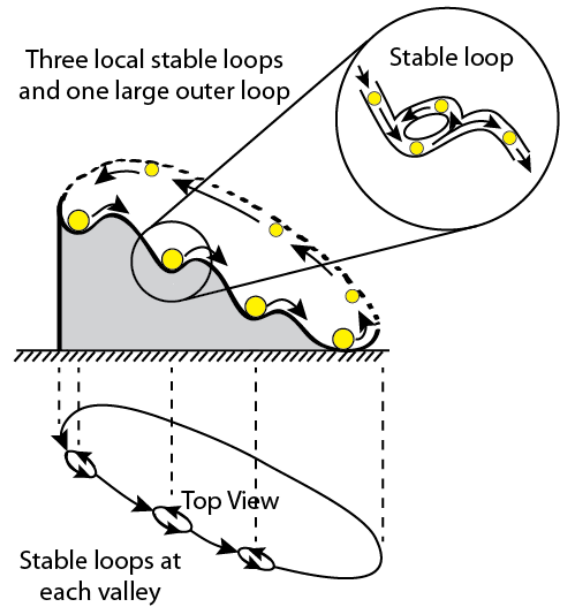

**Figure A1 | A simple set of physical SPL systems.** We consider a simple SPL system in which balls roll down an inclined plane while a human brings it back to complete the loop. We modify this system incrementally by eliminating the human and by adding more complexity. There are several ways to modify the system. I show one in which more locally stable points are added, which are then converted into locally stable loops.

To decrease the losses further, Fig. A1(d) shows another SPL design in which the stable points are replaced with tiny and well-lubricated stable loops. The balls entering the loop continue their looped dynamics in a single direction avoiding back-and-forth collisions. In addition, looped dynamics avoids the equivalent of high activation energy necessary to restart the motion, had each ball halted its motion. This lowers the dependency on the external world. The extra energy needed to kick the balls to the next higher level is not just from the local bottommost location as in Fig. A1(c), but from any other intermediate location as well, due to their continued motion within the looped path. If there is competition or a genuine shortage of energy, designs similar to Fig. A1(d) (like atomic orbital models with similar transitions between levels and molecular orbital dynamics) will exist much longer than the other designs.

From the above description, we can not only create a large family of physical SPL systems but we can also iteratively improve the operational inefficiencies of these designs. The generic process (Fig. A2(a)) is as follows: (i) introduce a stable loop which typically has one physical stable point, (ii) change the single physical stable point into multiple stable points, (iii) change one or more stable points into stable loops and (iv) continue step (i) for each stable loop iteratively. Fig. A2(a) shows the evolution of one SPL design after two iterations.

It is now possible to merge, link or chain (Fig. A2(b)) multiple SPL systems in an infinite number of ways with arbitrary complexity to create new SPL systems with an increased set of loops and an increased ability to exhibit interesting behaviors. During this process, we need to ensure that the parallel loops are coordinated to have a single direction of flow for achieving stability for a long time. Blood circulation systems and our brain networks are examples of similar physical SPL systems with directed flows.

Most classes of interconnected dynamical systems have the property that the higher the complexity, the easier it is to destabilize them. SPL systems are the only known class for which the opposite can be satisfied – higher complexity makes them easier to self-sustain longer (cf. Figs. A1(a) - A2). However, the above physical SPL systems cannot be created naturally. To circumvent this difficulty, we need to switch to chemical SPL systems.

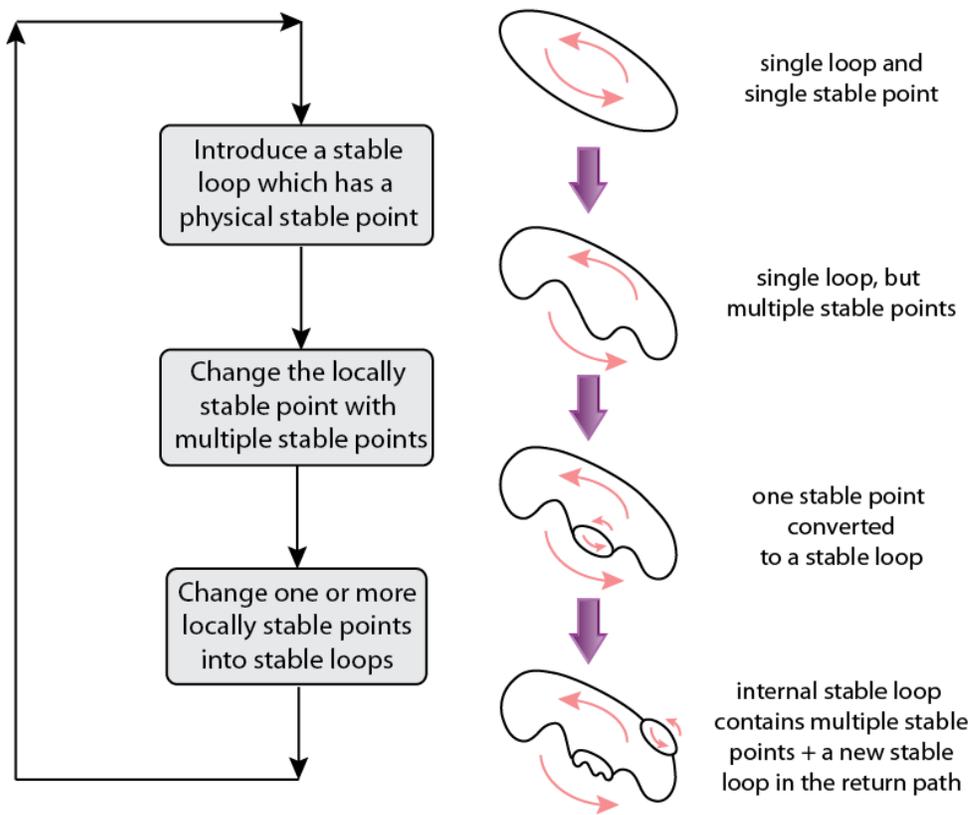

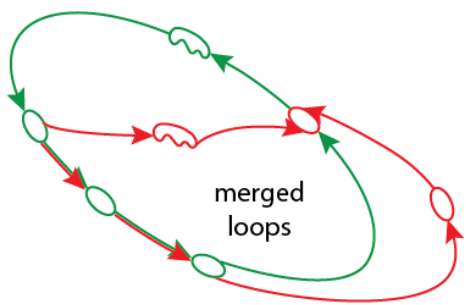
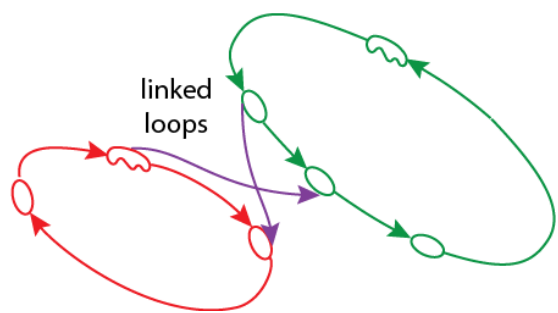

**Figure A2 | An iterative process to create an infinite family of physical SPL systems with increasing complexity.** Starting from a single looped SPL system, it is possible to create a multitude of physical SPL systems depending on how we modify subcomponents of it at various stages of evolution. We can either merge or link two or more physical SPL systems to create a new SPL system with different features and properties.

## A2.     Chemical SPL systems

Given the above infinite family of physical SPL systems, we can show the 'existence' of a large family of chemical SPL systems via a mapping specified in Fig. A3. Map all the balls (considered the same for physical SPL systems) as 'different' types of molecules. Represent different heights of the inclined plane as different energy levels. A ball at a given height is represented by a specific molecule with a corresponding energy state. The act of balls rolling downhill represents a chemical reaction in which reactant molecules of higher-energy state transition to product molecules of lower-energy state (Fig. A3). Though the balls in physical SPL systems do not change, the corresponding representation as molecules in a chemical SPL system do change with each transition. The valleys correspond to special locally stable molecules of a given energy level. The hills represent the activation energy that a chemical reaction needs to overcome.

There are multiple ways to map the static physical structure of inclined planes in the chemical world. In one representation, they correspond to dynamical enzymes (Fig. A3). Enzymes assist a given uphill and downhill chemical transition. While the balls directly absorb the external energy to produce the motion along the inclined plane, with chemical SPL systems, the enzymes absorb the energy instead. Their unique shape and size cause the uphill or downhill transition to create new products (cf. induced-fit model of enzymes). In another representation, they correspond to suitable conditions like temperatures, pressures and energies that allow a given chemical transition. Other representations include closed compartmental structures, catalysts, pumps and a combination of these. The operational aspects are similar with these representations, though the degree to which these structures assist the uphill and downhill reactions differ compared to enzymes.

Fig. A4 shows one example of chemical SPL system utilizing the above mapping. It contains two downhill reactions and one uphill reaction to complete a chemical loop. The chemical loop itself is virtual unlike a physical loop. The downhill reaction tends to occur easily and naturally while the uphill reaction requires considerable external energy and several favorable conditions, all of which are nontrivial to setup naturally, reliably and repeatedly. In Fig. A4, enzymes assist in converting molecules both in the downhill and uphill directions to form a simple chemical loop. An example of a complex organic chemical SPL system involving multiple loops is Kreb's cycle.

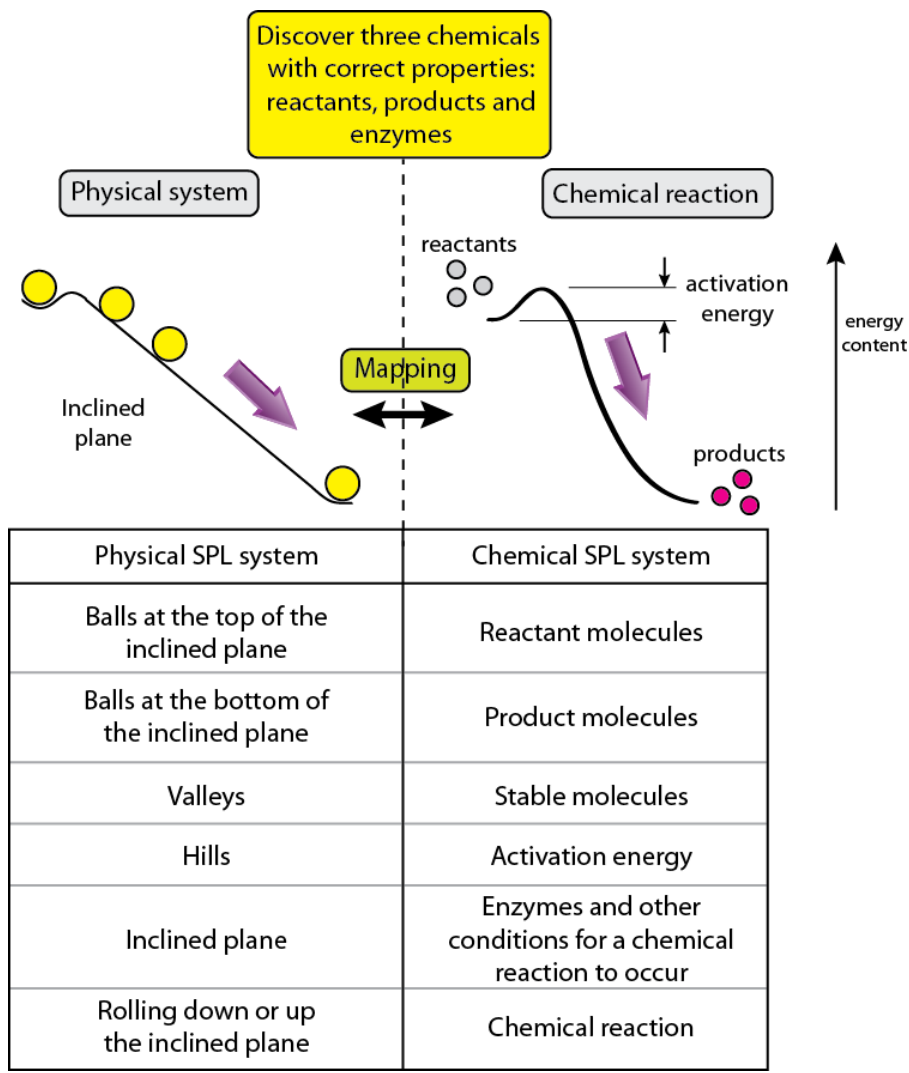

**Figure A3 | Mapping a physical SPL system to a chemical SPL system.** Balls rolling down an inclined plane from one energy state to another are mapped to an uphill or a downhill chemical reaction. As shown in the table, the activation energy, the enzymes and other conditions that assist a chemical reaction are mapped, provided reactants, products and enzymes with suitable properties exist or can be discovered.

From Fig. A3 and Fig. A4, it appears as though mapping a physical SPL system into a chemical SPL system is quite simple in an 'abstract' sense. However, creating a 'real' chemical SPL system is nontrivial because we need to discover different types of chemicals and enzymes with several relative constraints between them. As an example, consider Fig. A4. We need to represent three reactions for the mapping of the physical SPL system. From Fig. A3, each reaction requires three chemicals – reactants, products and enzymes. Therefore, we need to discover nine types of chemicals. However, since the physical and chemical reactions are interconnected, they impose constraints – the products of one reaction = reactants of another reaction. This implies we only need three chemicals ($M_1$, $M_2$ and $M_3$) and three enzymes ($E_{12}$, $E_{23}$ and $E_{31}$). Furthermore, the relative energy levels should also be such that $M_1 > M_2 > M_3$. Besides, the activation energies of each reaction (which are altered by enzymes) should be comparable to the corresponding depth of valleys in the physical SPL system that is being mapped.

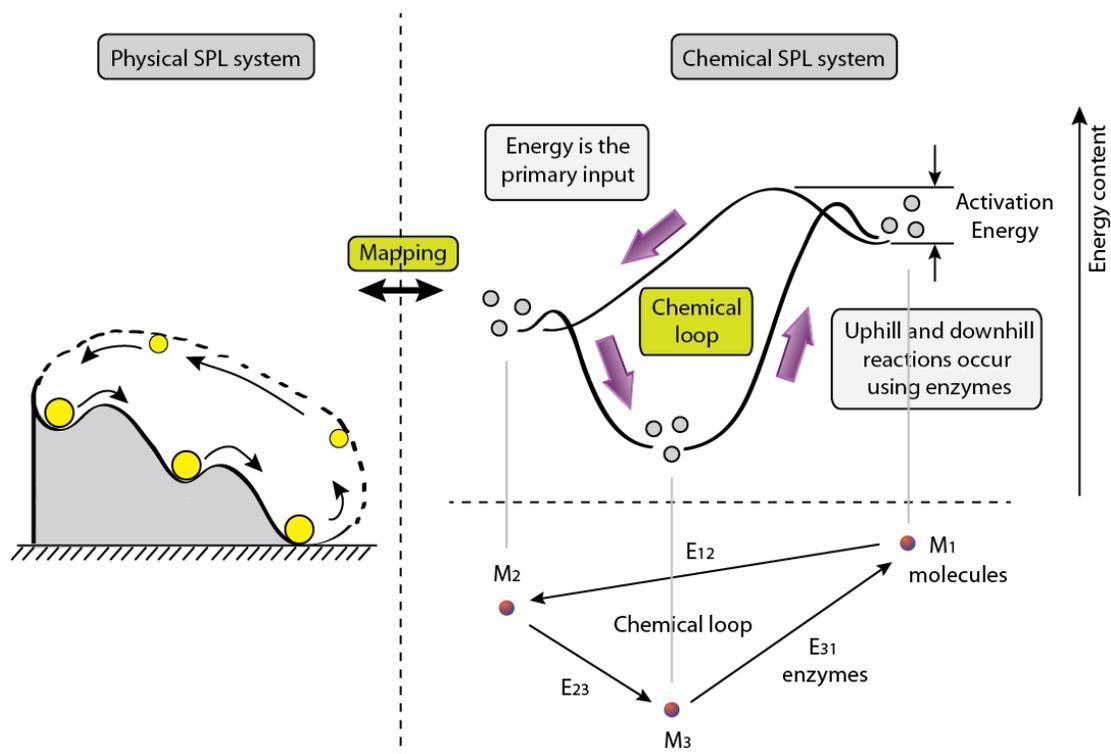

**Figure A4 | A sample chemical SPL system mapped from a physical SPL system.** Using the mapping of Fig. A3, a physical SPL system is mapped into a corresponding chemical SPL system. Most metabolic pathways within living beings (like the urea cycle and the Kreb's cycle) are much more complex than this sample chemical SPL system. Each reaction requires the discovery of a suitable enzyme both in the uphill and downhill directions to complete a loop.

For more complex physical SPL systems (like Fig. A2), the practical difficulty of discovering a real chemical SPL system that satisfies all constraints required for the correct mapping gets compounded. In this case, having a repository of metabolic networks from each species greatly helps us in finding chemicals with required relative relationships. Each design in Figs. A1-A2 can now be represented as chemical SPL systems using the above mapping with a suitable choice of chemicals and enzymes, if they exist.

Merging, linking, chaining and the generic process mentioned above for physical SPL systems continue to be applicable here (Fig. A5). They help generate a diverse and complex set of chemical SPL systems with an appropriate choice of chemicals like enzymes, organic reactants and products. As an example, the equivalent of replacing stable points (Fig. A3) with stable loops (Fig. A4) is glucose and $O_2$ forming $H_2O$ and $CO_2$ being replaced with a stable looped reaction, namely Kreb's cycle (Fig. A5(a)). In general, if we pick all the chemicals from the metabolic network of a given organism, we get a chemical SPL system. Therefore, *each living being can be viewed as a chemical SPL system*. Even though the above mapping is direct with organic reactions, we can, nevertheless, represent inorganic chemical loops like Belousov-Zhabotinsky and Briggs-Rauscher reactions as interacting SPL systems using the above mapping.

As we improve the SPL systems by building large-scale metabolic networks incrementally, the designs utilize the existing resources effectively and create, possibly, new chemicals to share across linked SPL systems. The set of waste products are iteratively minimized. An SPL system is considered *complete* if the only external dependencies are abundantly available chemicals and energy. For example, we would have to include Kreb's cycle within animals, Calvin-Benson cycle from plants, the chemical loops to create different enzymes and other metabolic pathways to create a complete SPL system within living beings, with sunlight as the primary external input. In a subsequent paper, I will show how such complex interconnected SPL systems exhibit several life-like behaviors, besides survival ability, seen simply as the ability to sustain the dynamics stably for a long time, discussed here (like image/speech recognition and others – Patent Pending).

While the question of existence of an infinite family of chemical SPL systems with any level of complexity is addressed with the above mapping, how do we address the creation of these systems naturally? The reason this is important is because *creating a living being and, hence, the first life forms from inanimate objects is equivalent to creating a suitable chemical SPL system* . From Fig. A5, creating a chemical SPL system require us to create a corresponding set of enzymes, other molecules of specific energy states and the setting up of conditions like energy, temperatures, pressures and closed compartments. After the origin of life, this task falls within the realm of Darwinian evolutionary processes. However, before the origin of life, there are no enzymes or DNA present.

Therefore, the objective with the paper is to discuss an alternate way to create chemical SPL systems naturally during this period avoiding enzymes and DNA. I will show that this will involve using special chemical pumps, which are by themselves naturally easy-to-create SPL systems. The pumps can be seen as the minimal structures (Ravuri 2011), necessary to create chemical SPL systems naturally. In this sense, the study of chemical SPL systems is closely related to the study of origin of life.

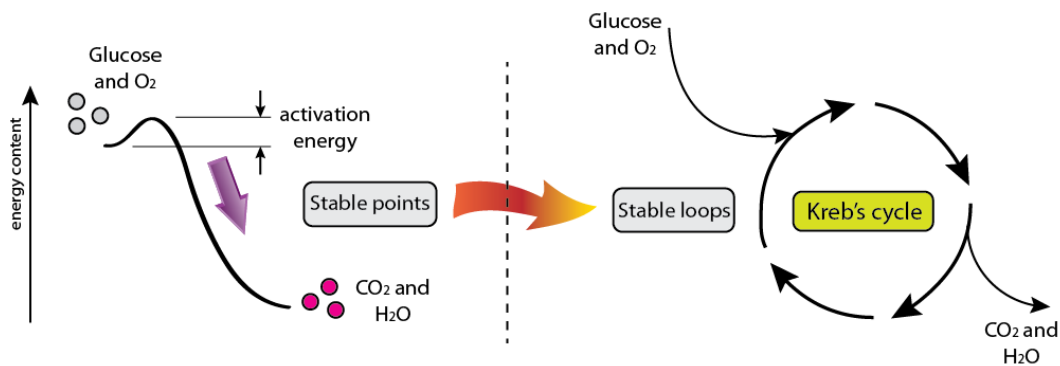

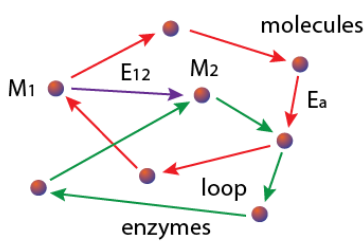
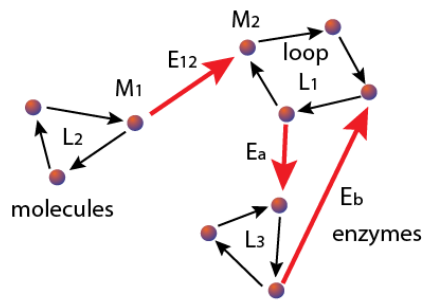

**Figure A5 | Mechanisms to create an infinite family of chemical SPL systems. a,** A chemical reaction with a stable point can be converted into a new SPL system with a chemical loop with the discovery of several new chemicals and enzymes like with Kreb's cycle. **b,** Two or more chemical SPL systems can be merged to create a new SPL system. **c,** Two or more chemical SPL systems can be linked to create a new SPL system. Using each of these processes, we create an infinite family of chemical SPL systems. Each of these processes requires the discovery of new enzymatic pathways.

## A3.  Self-replicating property

Chemical SPL systems exhibit self-replication property, which is a unique feature to life. This follows directly from Claim A1. If we have a collection of molecules and ask which subset remains after a long time, the typical answers are (a) inert, less reactive chemicals and ones with high half-life, (b) abundantly available chemicals and (c) chemicals that switch repeatedly and randomly between states. However, with the discovery of chemical SPL systems, we now have a new 'dynamical' case as well.

**Claim A1:** The subset of chemicals that will remain after a long time with high probability from any given collection of molecules are those that form a chemical SPL network, provided this network depends only on freely available external inputs/energy and molecules that are themselves part of other external loops.

*Proof*: From any collection of molecules, consider two subsets – one that forms chemical loops and another that does not. Molecules within the looped subset (a) reuses raw materials while consuming only abundant energy or chemicals unlike nonlooped subset, which produce non-reusable products and (b) self-sustain longer compared to nonlooped dynamics, which die down quickly. Additionally, nonlooped chemicals arriving within a small neighborhood at the correct time as coincidences so they could react, has near-zero probability (cf. air molecules reaching one corner of a room). Looped chemicals, on the other hand, are repetitively produced. As their concentrations keep increasing in a given region, the chances of coincidences improve. Therefore, the probability of looped subsets remaining after a long time is higher relative to nonlooped subsets. Q.E.D

If we view the static part of the chemical SPL network as a directed graph (see metabolic networks), we can partition it into a collection of strongly connected components. Each strongly connected component has at least one loop. Such a looped network is said to be generated by a subset $S$ of chemicals if and only if the entire network can be recreated using $S$ and the collection of enzymes. With the above partition, the subset $S$ needs to have only one chemical for each strongly connected component. Therefore, replicating the entire chemical SPL network is easy as there are several choices for subset $S$ of chemicals (for example, three strongly connected components with four chemicals in each has 64 choices). A simple random split of the chemical SPL system into two parts will regenerate the entire SPL network in each part. Even though no individual type of chemicals remains fixed or has long half-life, the entire SPL collection, as a whole, self-sustains for a long time. This self-replication property is different from DNA replication because it involves the entire looped collection to be replicated, not an individual molecule like a DNA sequence.

From Claim A1, the molecules that continue to exist for a long time are unrelated to the advantages they offer to the system, only that they are part of a looped network. The statement is applicable to both living and nonliving systems. One special case of Claim A1 in which the chemicals that remain after a long time are D-form sugars and L-form amino acids is the problem of homochirality. I will show how to achieve asymmetric distribution starting from a perfectly symmetric distribution of L- and D-forms as a direct consequence of chemical SPL structure in a subsequent paper. Claim A1 is true even if some of the reactions are catalyzed. The catalysts in this case would need to be freely available (like clay or iron) or should be created through other looped processes (like how enzymes are created from within metabolic networks).

**Claim A2:** Even though no individual type of chemicals remains fixed or has long half-life, the entire SPL collection, as a whole, self-sustains for a long time.

This is the *self-reproduction* property of a connected chemical SPL network – starting with an initial threshold but random collection of chemicals we can regenerate the rest of the looped network under reasonable conditions. For example, a simple random split into two parts will regenerate the entire SPL network in each part. Nonlooped networks do not have this property. This is because when a cascading linear chain of reactions occur they decrease the concentrations of intermediate chemicals thereby breaking the continuity of the chain. Looped reactions, on the other hand, regenerate and maintain steady concentrations of all intermediate chemicals thereby guaranteeing causal continuity across a long chain of reactions. This self-replication property is different from DNA replication because it involves the entire looped collection to be replicated, not individual molecule like a DNA sequence.

Claim A2 lets us move away from, say, discovering RNA molecules with the property of catalyzing the replication of RNA from an RNA template (i.e., an RNA replicase) individually and instead focus on discovering an entire SPL collection within which one of the RNA sequence does exhibit the above special property.

One inherent assumption in the above discussion with the sustainability of the dynamics of chemical SPL systems is that the rate of the looped reactions, which maintains a steady production of chemicals, should be faster than the effects of second law, which tries to breakdown the system. For example, volcanoes, meteor strikes and other unexpected second law effects should not collapse the SPL system faster than it can recover. Even otherwise, when better conditions return, looped reactions will restart with higher probability compared to nonlooped collections.

## Acknowledgements

I thank my wife Sridevi Ravuri for over ten years of discussions, comments and for the emotional support.